\documentclass[runningheads]{llncs}
\usepackage[T1]{fontenc}
\usepackage{graphicx}
\usepackage{makeidx}  
\usepackage[colorlinks,linkcolor=blue]{hyperref}
\usepackage{url}
\usepackage{longtable}
\usepackage{multirow}
\usepackage{amssymb, amsmath, bm}
\usepackage{amsfonts, amssymb}
\usepackage{pifont}
\usepackage{mathtools}
\usepackage{mathrsfs}
\usepackage{booktabs}
\usepackage{cite}
\usepackage{dsfont}
\usepackage{makecell}
\usepackage{multirow}
\usepackage[noend]{algpseudocode}
\usepackage{algorithmicx,algorithm}
\usepackage[misc]{ifsym} 
\usepackage{bbding}
\usepackage{ulem}
\begin{document}
\title{
Mitral Regurgitation Recognition based on Unsupervised Out-of-Distribution Detection with Residual Diffusion Amplification}
\titlerunning{Residual Diffusion Amplification}
%
%
\author{
Zhe Liu\inst{1,2}\thanks{Zhe Liu, Xiliang Zhu and Tong Han contribute equally to this work.\\
Corresponding authors’ email: \email{gzhongshan1986@163.com} and  \email{xinyang@szu.edu.cn}}\and
Xiliang Zhu\inst{1,2\star}\and
Tong Han\inst{1,2\star}\and 
Yuhao Huang\inst{1,2}\and 
Jian Wang\inst{3}\and\\
Lian Liu\inst{1,2,4}\and
Fang Wang\inst{5}\and
Dong Ni\inst{1,2}\and
Zhongshan Gou\inst{5}\textsuperscript{(\Letter)}\and\\
Xin Yang\inst{1,2}\textsuperscript{(\Letter)}
}

\authorrunning{Liu et al.}
%
\institute{\textsuperscript{$1$}National-Regional Key Technology Engineering Laboratory for Medical Ultrasound, School of Biomedical Engineering, Medical School,
Shenzhen University, China\\
\textsuperscript{$2$}Medical Ultrasound Image Computing (MUSIC) Lab, Shenzhen University, China\\
\textsuperscript{$3$}Key Laboratory for Bio-Electromagnetic Environment and Advanced Medical Theranostics, School of Biomedical Engineering and Informatics,\\ Nanjing Medical University, China\\
\textsuperscript{$4$}Shenzhen RayShape Medical Technology Co., Ltd, China\\
\textsuperscript{$5$}Center for Cardiovascular Disease, The Affiliated Suzhou Hospital of Nanjing Medical University, China\\
}
\maketitle              
\begin{abstract}
Mitral regurgitation (MR) is a serious heart valve disease. 
Early and accurate diagnosis of MR via ultrasound video is critical for timely clinical decision-making and surgical intervention. 
However, manual MR diagnosis heavily relies on the operator's experience, which may cause misdiagnosis and inter-observer variability. 
Since MR data is limited and has large intra-class variability, we propose an unsupervised out-of-distribution (OOD) detection method to identify MR rather than building a deep classifier.
To our knowledge, we are the first to explore OOD in MR ultrasound videos.
Our method consists of a feature extractor, a feature reconstruction model, and a residual accumulation amplification algorithm. 
The feature extractor obtains features from the video clips and feeds them into the feature reconstruction model to restore the original features.
The residual accumulation amplification algorithm then iteratively performs noise feature reconstruction, amplifying the reconstructed error of OOD features.
This algorithm is straightforward yet efficient and can seamlessly integrate as a plug-and-play component in reconstruction-based OOD detection methods.
We validated the proposed method on a large ultrasound dataset containing 893 non-MR and 267 MR videos. 
Experimental results show that our OOD detection method can effectively identify MR samples.

\end{abstract}
\section{Introduction}
\begin{figure}[!h]
\centering
\includegraphics[width=1.0\linewidth]{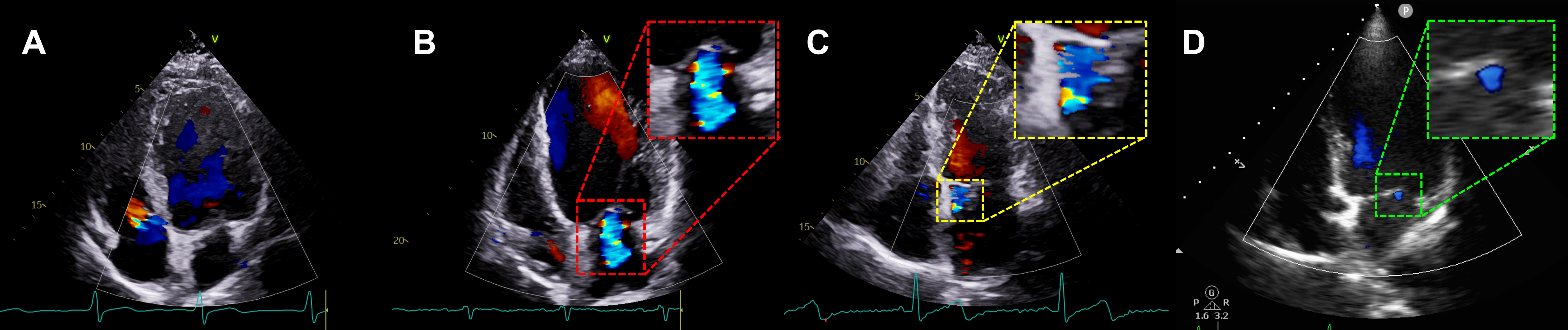}
\caption{The challenges encountered in OOD detection for our task.~\textbf{A)} Negative sample,~\textbf{B)} Positive sample,~\textbf{C)} Negative sample with artifacts similar to regurgitation,~\textbf{D)} Positive samples of reflux bundles resembling artifacts.} \label{fig:OOD_data}
\end{figure}

Mitral regurgitation (MR) is a significant heart valve disorder that becomes more common with advancing age.
It may cause heart size changes, decreased cardiac function, and even life-threatening consequences~\cite{el2018mitral}.
Early detection and accurate assessment of MR enables timely intervention, holding vital clinical significance.
Color Doppler echocardiography is the primary tool to diagnose MR, enabling visualization of blood flow direction and velocity within the heart that is not visible in standard B-mode ultrasound~\cite{zoghbi2003recommendations}. 
However, manual assessment is subjective and easily affected by experts' experience, potentially leading to diagnostic errors and observer inconsistency.  
Figure~\ref{fig:OOD_data} shows the challenges in MR recognition, where similar features appear in both negative and positive cases.

Deep learning techniques can potentially address the above problems, e.g., by training a classifier to recognize MR. 
However, the limited availability of MR data and its significant intra-class variability compared to normal data can constrain the performance of binary classifier~\cite{lu2015simplified}. For example, the imbalanced data distribution caused by the lack of MR data can severely affect the performance of supervised methods~\cite{hendrycks2016baseline,zhang2023mixture,mishra2023dual}. 
Recently, there have been significant advancements in out-of-distribution (OOD) detection, opening up new possibilities for MR recognition.
The goal of OOD detection is to identify samples that are inconsistent with the distribution of in-distribution (ID) data (i.e., training data)~\cite{yang2021generalized}. 
It is noted that in MR recognition, numerous normal samples with similar features are regarded as ID data, and minimal but diverse MR cases are considered OOD data. 
Although promising, we found that studies applying OOD detection to MR recognition have not yet been reported. 

Applying existing OOD research to MR recognition without modification may face several challenges. 
Unsupervised OOD detection methods for videos showed great potential, and they mainly included two main streams: frame prediction and generation techniques.
Among frame prediction methods, those using optical flow maps~\cite{duman2019anomaly,baradaran2022object} and long sequences of observations~\cite{baradaran2023future,wang2021robust} were deemed inadequate for MR ultrasound videos due to noise, artifacts, and the cardiac cycle's motion pattern.
Besides, generation-based approaches may encounter information bottlenecks caused by discrepancies between potential input dimensions and the dimensions reconstructed by the model~\cite{graham2023denoising,serra2019input}.
The diffusion model solely employed the mean and standard deviation for denoising the reconstructed image, thereby mitigating the bottleneck issue mentioned previously to ensure the reconstruction of echocardiographic videos~\cite{wyatt2022anoddpm,graham2023denoising,mishra2023dual}.

In this study, we propose a diffusion-based unsupervised OOD detection method for recognizing MR from echocardiography videos. 
Specifically, we first employ pre-trained models as feature extractors to extract features from video clips. 
Second, we train a diffusion model for feature-level reconstruction. 
Last, we design a residual accumulation amplification algorithm during the testing phase.
This algorithm iteratively performs noise feature selection and reconstruction, amplifying the reconstructed error of OOD features.
We validate the proposed method on a large four-chamber cardiac (4CC) ultrasound video dataset containing 893 non-MR and 267 MR videos, and the experimental results show that the proposed method is effective. 
We believe we are the first to explore OOD detection in MR recognition with echocardiography videos.


	





\section{Methodology}
\begin{figure}[!h]
\centering
\includegraphics[width=1.0\linewidth]{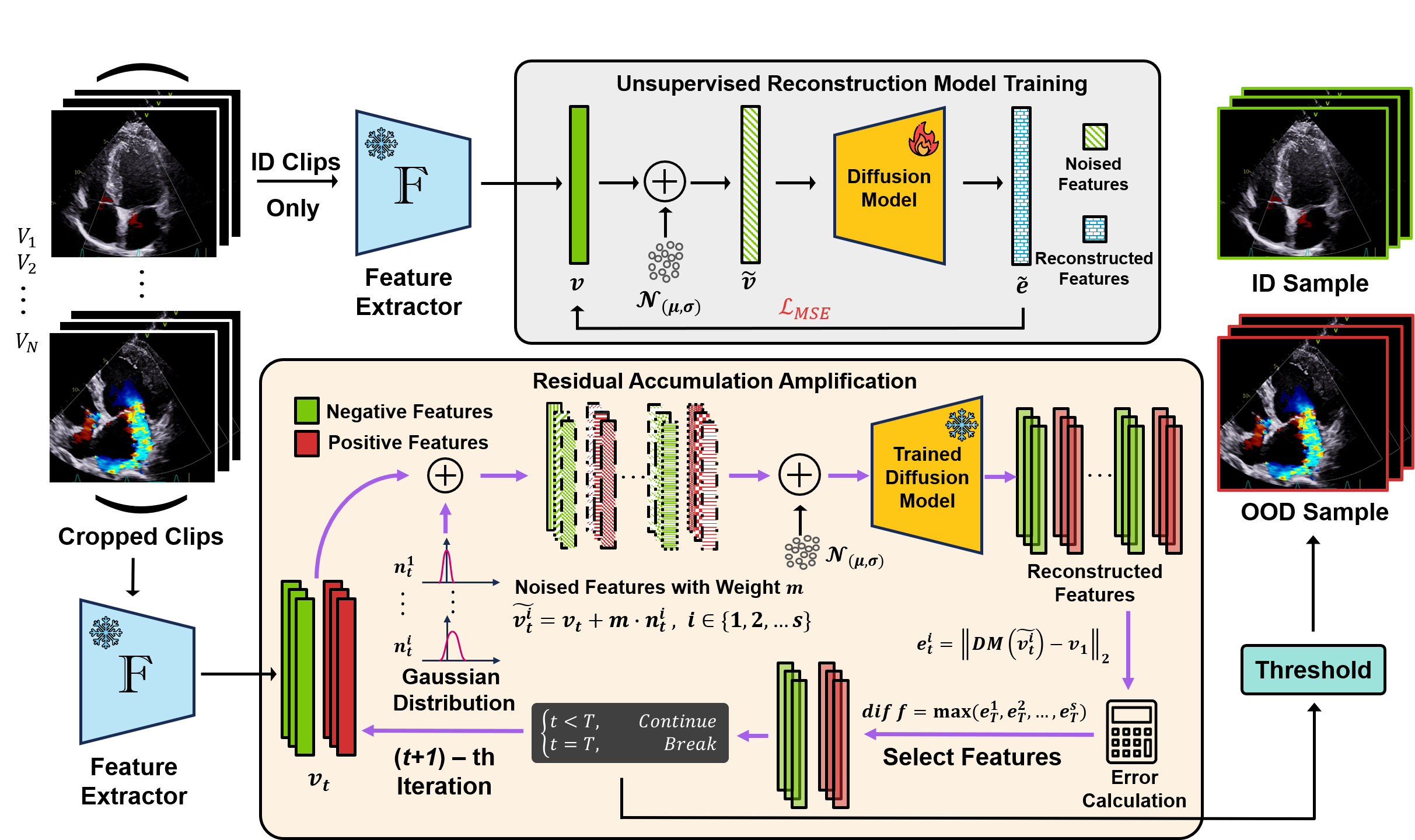}
\caption{Overview of our proposed method.} \label{fig:network}
\end{figure}

As shown in Figure~\ref{fig:network}, our method consists of a feature extractor, a feature reconstruction model, and a residual accumulation amplification algorithm. 
The feature extractor catches features from video clips and inputs them into the reconstruction model to rebuild the features. 
Then, during testing, under the proposed residual accumulation amplification algorithm, the OOD data can gradually be more distinguishable from the ID data.

\subsection{Representative Feature Extractor}
The feature extractor aims to capture essential knowledge from video clips. Let $V \in R^{k \times 3 \times h \times w}$ denote a video clip consisting of \textit{k} consecutive RGB images with a resolution of $h \times w $ pixels, the feature $v \in R^{l}$ is obtained by feeding $V$ into the feature extractor $f(\cdot)$. 
The architecture of the feature extractor is flexible and can be convolutional neural networks, long short-term memory, visual transformers, and their hybrids. 
Here, we tried several models, including 3D-ResNet18~\cite{hara2017learning}, 3D-ResNet101~\cite{hara2017learning}, Video MAE~\cite{tong2022videomae}, and X-CLIP~\cite{ni2022expanding}. 
These models were pretrained on Kinetics-400~\cite{carreira2017quo}, a large-scale dataset containing more than 160,000 videos with 400 categories. We froze the models' weights without any fine-tuning.

\subsection{Feature Reconstruction Model}
The diffusion model gains the capability to generate diverse samples by disrupting training samples with noise and learning the reverse process.
For a given video clip $V$, features are obtained by $v=f(V)$.
The process of feature disruption can be formulated as a progressive addition of Gaussian noise (standard deviation: $\sigma$) to the input data $v_T$ obtained by sampling from a data distribution $p_{data}(v)$ with standard deviation $\sigma_{data}$. After adding noise, the data distribution could be represented as $p(v;\sigma)$ and $p\left(v;\sigma_{min}\right)\approx p_{data}(v)$ ideally. If $\sigma_{max}\gg\sigma_{data}$, the $p\left(v;\sigma\right)$ becomes isotropic Gaussian and allows to sample a data $v_0\sim \mathcal N(0,\sigma_{max}I)$. The reversed process of diffusion model can be formulated as $v_0$ being gradually denoised with noise levels $\sigma_0=\sigma_{max}>\sigma_1>\ldots>\sigma_T=0$ into new samples, until it becomes a data with data distribution of $p_{data}(v)$. 




In detail, we follow the diffusion model in~\cite{Karras2022Advances}, treating ${s_\theta}(\widetilde v,\sigma)$ as a denoising function that directly estimates denoised samples with target function $\mathcal L$:
\begin{equation}
\mathcal L=E_{v\sim p_{data}}E_{\epsilon  \sim \mathcal N(0,\sigma I)}\Vert s_\theta\left(v+\epsilon;\sigma\right)- v \Vert_{2}.
  \end{equation}

Besides, in~\cite{Karras2022Advances}, researchers untangled the design choices of previous diffusion models and offered a framework where each component (Sampling, Network and preconditioning, Training) can be adjusted independently. Therefore, in our study, the denoising function $s_{\theta}(\cdot )$ is formulated as follows:
\begin{equation}
s_{\theta}(v;\sigma)=c_{skip}(\sigma)v+c_{out}(\sigma)*F_{\theta} (c_{in}(\sigma)v;c_{noise}(\sigma)),
  \end{equation}
where $F_{\theta}(\cdot)$ is a multi-layer perceptron (MLP) with an encoder-decoder structure. 
$c_{skip}(\cdot)$ modulates the skip connection.
$c_{out}(\cdot)$ and $c_{in}(\cdot)$ scale magnitudes of the noise variance.
$c_{noise}(\cdot)$ scales $\theta$ to a suitable value for the input of $F_{\theta}(\cdot)$.

The reverse process of disrupting samples in the feature reconstruction of video clips can be formulated as:
\begin{equation}
\widetilde v_{t}=\widetilde v_{t-1}+\frac{\varepsilon }{2} s_{\theta }(\widetilde v_{t-1}, \sigma )+\sqrt{\varepsilon}z_t,
  \end{equation}
where { $\varepsilon$ > 0} is a predefined step size and  $z_t\sim \mathcal N(0,I)$ is  random term. With certain constraints, reversing multiple times at very small step sizes results in the final generated sample obeying the distribution $p_{data}(v)$.

The diffusion process is governed by multiple hyperparameters, with our main focus being the exploration of the effect of noise distribution on model performance. 
For fair comparisons, we follow~\cite{Tur2023Unsupervised} to determine other settings, including model structure, learning rate scheduler, etc.
The diffusion model is central to the OOD detection model, used to recover the samples after being corrupted with Gaussian noise during the testing phase. 
The OOD information is derived by evaluating the discrepancy between samples before and after restoration.

\subsection{Residual Accumulation Amplification}
Inspired by the genetic accumulation of mutated genes with inheritance in genetics~\cite{holland1992adaptation}, we designed a residual accumulation amplification algorithm used in the testing stage. 
The core of the algorithm involves selecting MR recognition-related features and amplifying the reconstruction error via multiple iterations of noise feature reconstruction to enhance the distinction between OOD and ID data.
In each iteration, we first randomly sample $s$ Gaussian noises from the normal distribution $\mathcal N(\mu,\sigma)$, and then add them to the anchor feature to randomly destroy features that represent different information. The formula is as follows:
\begin{equation}
  \widetilde v_t^i = v_t + m \cdot n_t^i, \quad i \in \left \{ 1,2,...s \right \},
  \end{equation}
where $n^i$ denotes the sampled Gaussian noise and \textit{m} denotes the weight of noise. $v$ and $\widetilde v^i$ represent the anchor and noise features, respectively. $t$ is the current iteration number. 
In the first iteration, the anchor feature ($v_1$) is extracted from the video clips using the feature extractor. 
Those noise features are fed into the reconstruction model, and the reconstructed error is calculated as follows:
\begin{equation}
e_t^i = \Vert DM(\widetilde v_t^i) - v_1 \Vert_2, \quad i \in \left \{ 1,2,...s \right \},
\end{equation}
where $DM(\cdot)$ denotes the reconstruction model and $e^i_t$ denotes the reconstructed error in the \textit{t}-th iteration. $\Vert \cdot \Vert_2$ represents the $L_2$ norm. 

As shown in Figure~\ref{fig:point} (D), OOD data is more sensitive to reconstruction, and the difference between the reconstruction error of ID and OOD data will increase after multiple reconstructions.
Therefore, we continue to add Gaussian noise to the current noise features for the next reconstruction. 
The noise feature corresponding to the maximum error is selected as the anchor feature in the next iteration. We believe that the larger error arises from the destruction of features unrelated to MR recognition. 
As these irrelevant features are eliminated after multiple iterative selections, the remaining features that are more related to MR recognition can enhance the detection efficacy: 
\begin{equation}
v_{t+1} = \mathop{\arg\max}\limits_{\widetilde v_t}(e_t).
\end{equation}

The reconstruction error continues to be amplified until the stopping condition is reached, that is, $\textit{t} = \textit{T}$. The maximum error in the last iteration is regarded as the difficulty of feature reconstruction: $diff = max(e_T^1, e_T^2, ... e_T^s).$
The sample is classified according to a data-driven threshold ($thre$), defined as $\textit{thre}=\mu_{diff} + 0.001 \sigma_{diff}$, where $\mu_{diff}$ and $\sigma_{diff}$ are the mean and standard deviation of the difficulty of feature reconstruction in the validation set.


\section{Experimental Results}
\subsection{Dataset and Implementations}
The 4CC ultrasound video dataset was collected, comprising 893 non-MR and 267 MR videos.
Regions of interest were extracted from frames originally sized at 1016$\times$708 and resized to 224$\times$224 for further processing.
Multiple non-overlapping video clips were sampled from each video, and each clip consists of 16 consecutive frames, resulting in 3480 clips. 
Those clips cropped from MR videos are designated as positive samples (i.e., OOD samples), while others are taken as negative samples (i.e., ID samples). 
The dataset was divided into a training set with 1506 ID clips, a validation set with 621 clips (372 ID/249 OOD), and a testing set with 1353 clips (801 ID/552 OOD).

We adopted log-normal sampling (mean: -0.05, var: 1.5) for noise generation. Training epoch is set to 100 with \textit{batch size}=64, using the AdamW optimizer with \textit{learning rate}=0.0002. We used five metrics for evaluation: F1-score, Recall, Precision, Specificity, and Accuracy. All experiments were conducted on the \textit{PyTorch} 2.0.1 framework with NVIDIA Tesla A40 GPUs.


\subsection{Performance Comparisons}
We compared our proposed method with a basic supervised classification method (3D-ResNet50) and other reconstruction-based unsupervised OOD detection methods under the same settings. 
In 3D-ResNet50, we gradually increased the number of OOD clips $N$ in the training set.
As shown in Table~\ref{tab:comparison of methods}, the proposed method surpasses previous unsupervised methods and even surpasses 3D-ResNet50 with limited OOD data.


\begin{table}[!t]
  \centering
      \small
  \caption{Comparison of our method with 3D-ResNet50 and unsupervised methods.}
  \resizebox{\textwidth}{!}{
    \begin{tabular}{c|ccccc}
    \Xhline{1.0px}
    Supervised Methods & \multicolumn{1}{l}{F1-score} & \multicolumn{1}{l}{Recall} & \multicolumn{1}{l}{Precision} & \multicolumn{1}{l}{Specificity} & \multicolumn{1}{l}{Accuracy} \\
    \hline
    3D-ResNet50(N=240)~\cite{hara2017learning} & 0.0246 & 0.0128 & 0.3333 & \textbf{0.9845} & 0.6167\\
    3D-ResNet50(N=360)~\cite{hara2017learning} & 0.7338 & 0.8213 & 0.6632 & 0.7461 & 0.7746 \\
    3D-ResNet50(N=480)~\cite{hara2017learning} & \textbf{0.7549} & \textbf{0.8255} & \textbf{0.6953} & 0.7798 & \textbf{0.7971}\\
    \hline
    Unsupervised Methods & \multicolumn{1}{l}{F1-score} & \multicolumn{1}{l}{Recall} & \multicolumn{1}{l}{Precision} & \multicolumn{1}{l}{Specificity} & \multicolumn{1}{l}{Accuracy} \\
    \hline
    Diffusion + Star~\cite{Tur2023Unsupervised} & 0.4921 & 0.5545 & 0.4423 & 0.5822 & 0.5719 \\
    Diffusion + Dyn~\cite{Tur2023Unsupervised} & 0.4964 & 0.5446 & 0.4561 & 0.6118 & 0.5867 \\
    HF$^2$VAD~\cite{liu2021hybrid} & 0.4800 & 0.5347 & 0.4355 & 0.5858 & 0.5667 \\
    Ours  & \textbf{0.5545} & \textbf{0.5883} & \textbf{0.5244} & \textbf{0.6342} & \textbf{0.6156}\\
    \Xhline{1.0px}
    \end{tabular}}%
  \label{tab:comparison of methods}%
\end{table}%

\subsection{Comparison of Feature Extractor}
We compared the MR recognition performance under different feature extractors.
Thus, we used X-CLIP as the backbone in the following experiments for it achieved the best results in Table~\ref{models_compare}.
The X-CLIP extractor with Transformer modules pretrained on a vast video dataset has a stronger capability to capture rich semantic information. 
This may be because text-video model fusion allows a more comprehensive understanding of complex features by capturing diverse aspects of the data. 
The performance gap between Convolution-based models and Transformer-based models can be attributed to the superior ability of Transformer with self-attention modules in learning long-range dependence features compared to convolutional modules.
Particularly in video tasks with temporal information, Transformers exhibit advantages due to their enhanced capability to capture temporal dependencies and patterns.

\begin{table}[!t]
  \centering
  \small
  \fontsize{11}{12}\selectfont
  \caption{Performance comparison of OOD models with different feature extractors.}
    \begin{tabular}{c|ccccc}
    \Xhline{1.0px}
    Model & F1-score & Recall & Precision & Specificity & Accuracy \\
    \hline
    3D-ResNet18~\cite{hara2017learning} & 0.3193 & 0.255 & 0.4268 & 0.7653 & 0.5578\\
    3D-ResNes101~\cite{hara2017learning} & 0.3218 & 0.2459 & 0.4655 & 0.8065 & 0.5785 \\
    VideoMAE~\cite{tong2022videomae} & 0.4124 & 0.4226 & 0.4028 & 0.5705 & 0.5104 \\
    X-CLIP~\cite{ni2022expanding} & \textbf{0.5116} & \textbf{0.5228} & \textbf{0.5009} & \textbf{0.6429} & \textbf{0.5941}\\
    \Xhline{1.0px}
    \end{tabular}%
  \label{models_compare}%
\end{table}%

\begin{figure}[!t]
\centering
\small
\includegraphics[width=1.0\linewidth]{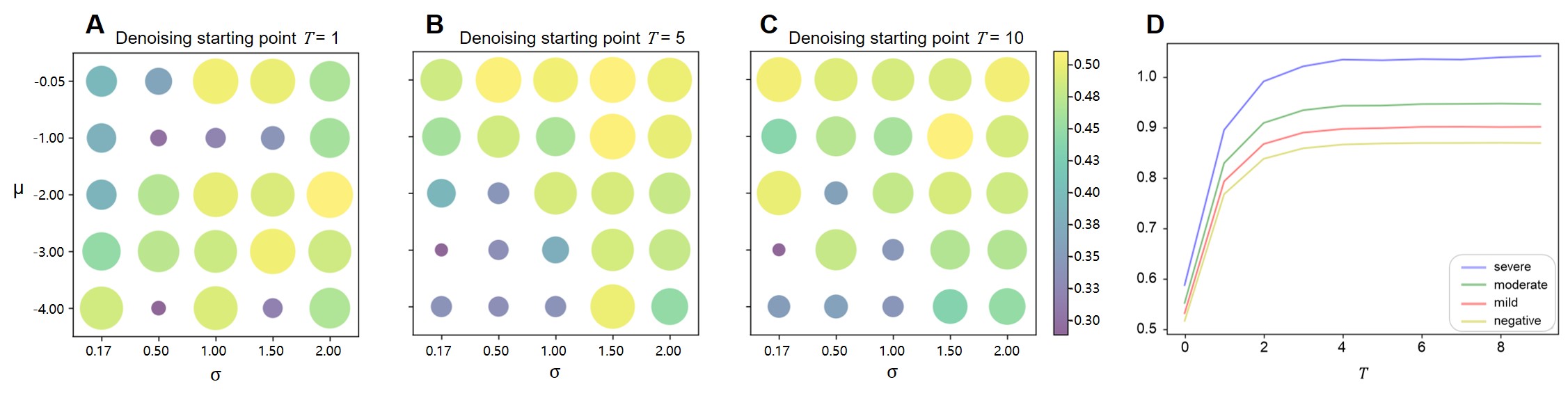}
\caption{\textbf{(A-C)} The effect of noise and the starting point of the reverse process. The larger the circle area, the greater the difference between the features. ~\textbf{(D)} Reconstruction difference among different subclasses of MR samples with feature amplification.
}
\label{fig:point}
\end{figure}

\subsection{Diffusion Model Analysis} 
We tend to find the optimal $T$ and $\mathcal N(\mu,\sigma)$ to preserve sufficient structural information of the video clip while disrupting potential anomaly information. 
Then, higher reconstruction errors can be obtained to judge the associated video frames as abnormal. 
For Gaussian distributions $\mathcal N(\mu,\sigma)$, $\mu$ ranges from -0.05 to -4 and $\sigma$ ranges from 0 to 2.
Besides, the start points $T$ of the reverse process are set to 1,5 and 10. 
For each value of $T$, 25 different Gaussian noise samples can be produced, as shown in Figure~\ref{fig:point} (A-C). It can be observed that $\mu=-0.05,~\sigma=1.5$ and $T=5$ obtain the best score. 
Overall, increasing the $T$ value while fixing $\mu$ and $\sigma$ can improve the OOD detection results. 
$\sigma$ is the primary determinant of the model. 
Specifically, the performance gap for changing the value of $\sigma$ is 3\% in the F1-score, with all other hyperparameters remaining constant.

\begin{table}[!t]
  \centering
    \small
  \caption{Performance Comparison with different parameters. Bold: the final model adopted in our work. The feature extractor is a pretrained X-CLIP.}
\setlength{\tabcolsep}{2.0mm}
\resizebox{\textwidth}{!}{
    \begin{tabular}{ccc|ccccc}
    \Xhline{1.0px}
    {\textit{T}}  & {\textit{s}}     & {\textit{m}}     & \multicolumn{1}{l}{F1-score} & \multicolumn{1}{l}{Recall} & \multicolumn{1}{l}{Precision} & \multicolumn{1}{l}{Specificity} & \multicolumn{1}{l}{Accuracy}\\
    \hline
    1     & 1     & 0     & 0.5116  & 0.5228  & 0.5009  & 0.6429  & 0.5941  \\
    2     & 1     & 0     & 0.5081  & 0.5118  & 0.5045  & 0.6554  & 0.5970  \\
    3     & 1     & 0     & 0.5036  & 0.5064  & 0.5009  & 0.6542  & 0.5941  \\
    4     & 1     & 0     & 0.5041  & 0.5046  & 0.5036  & 0.6592  & 0.5963  \\
    5     & 1     & 0     & 0.5246  & 0.5337  & 0.5158  & 0.6567  & 0.6067  \\
    10    & 1     & 0     & 0.5132  & 0.5118  & 0.5147  & 0.6650  & 0.6022  \\
    5     & 3     & 1     & 0.5426  & 0.5683  & 0.5191  & 0.6392  & 0.6104  \\
    5     & 3     & 1.5   & 0.5530  & 0.5847  & 0.5245  & 0.6367  & 0.6156  \\
    5     & 3     & 1.6   & 0.5547  & 0.5865  & 0.5261  & 0.6380  & 0.6170  \\
    5     & 3     & 1.7   & 0.5547  & 0.5865  & 0.5261  & 0.6380  & 0.6170  \\
    5 & 3 & 1.8 & \textbf{0.5545 } & \textbf{0.5883 } & \textbf{0.5244 } & \textbf{0.6342 } & \textbf{0.6156 } \\
    5     & 3     & 1.9   & 0.5515  & 0.5847  & 0.5220  & 0.6330  & 0.6133  \\
    5     & 3     & 2     & 0.5526  & 0.5883  & 0.5210  & 0.6292  & 0.6126 \\
    \Xhline{1.0px}
    \end{tabular}
    }%
  \label{parameters_compare}%
\end{table}%

\begin{table}[!t]
  \centering
  \small
  \fontsize{11}{12}\selectfont
    \caption{OOD performance of the reconstruction model with different modules.}
    \setlength{\tabcolsep}{0.8mm}
    \begin{tabular}{c|ccccc}
    \Xhline{1.0px}
    Model & F1-score & Recall & Precision & Specificity & Accuracy\\
    \hline
    Baseline & 0.5116 & 0.5228 & 0.5009 & 0.6429 & 0.5941\\
    Baseline+RA & 0.5246 & 0.5337 & 0.5158 & \textbf{0.6567} & 0.6067 \\
    Baseline+RA+NS & \textbf{0.5545} & \textbf{0.5883} & \textbf{0.5244} & 0.6342 & \textbf{0.6156}\\
    \Xhline{1.0px}
    \end{tabular}%
  \label{modle_comparison}%
\end{table}%

\subsection{Residual Accumulation Amplification Analysis}
Residual accumulation amplification consists of residual accumulation (RA) and noise sampling (NS). 
This section analyzes the impact of both components on MR recognition performance.
For RA, the reconstruction is iterated $T$ times (i.e., 1, 2, 3, 4, 5, and 10). Table~\ref{parameters_compare} shows that when $T$ is set to 5, the diffusion model achieves optimal detection results on all metrics, and $T$=10 causes a slight performance drop. 
For NS, we use the control variable method to assess how different parameters affect diffusion model performance.
First, we set the noise sampling number $s$=3 as a trade-off for computational efficiency. Then we change the weight of noise $m$ to analyze its impact. 
Results in Table~\ref{parameters_compare} show that consistent optimal results across all metrics can not be obtained with a single set of parameters.
But in general, the number of iterations $T$=5 can get good results easier. 
Finally, we choose the optimal hyperparameters ($m=1.8$, $s=3$, $T=5$) to construct the OOD detection model. 
The impact of each module on model performance is shown in Table~\ref{modle_comparison}.
RA improves the model performance by amplifing the difference between ID and OOD data reconstruction errors, which is consistent with Figure~\ref{fig:point} (D). However, its impact is limited by irrelevant features in the extracted information. NS addresses this issue effectively, significantly boosting performance by retaining MR recognition-relevant features through iterative selections.

\section{Conclusion}
We propose an innovative model tailored for recognizing MR echocardiographic videos using an unsupervised OOD detection approach. 
Our model exclusively leverages features extracted from ID samples by a fixed-weight feature extractor to train the diffusion model. 
Subsequently, by iteratively introducing disturbance, the model selects MR recognition-related features and amplifies the difference between reconstruction errors of OOD and ID data. 
Comprehensive experimentation conducted on MR datasets substantiates the efficacy of our proposed method. 
Future endeavors may explore expanding our approach to encompass multi-classification within the OOD samples.
\begin{credits}
\subsubsection{Acknowledgement.}
This work was supported by the grant from National Natural Science Foundation of China (12326619, 62101343, 62171290), Science and Technology Planning Project of Guangdong Province (2023A0505020002), Shenzhen-Hong Kong Joint Research Program (SGDX20201103095613036) and Suzhou Gusu Health Talent Program (GSWS 2022071, GSWS 2022072).

\subsubsection{Disclosure of Interests.}
The authors have no competing interests to declare that are relevant to the content of this article.
\end{credits}

%
%
%
\bibliographystyle{splncs04}
\bibliography{reference}
\end{document}